\documentclass[10pt,twocolumn,letterpaper]{article}

\usepackage{cvpr}
\usepackage{pgfplots}
\usepackage{times}
\usepackage{epsfig}
\usepackage{graphicx}
\usepackage{amsmath}
\usepackage{amssymb}
\usepackage{soul}
\usepackage{subcaption}
\usepackage{comment}
\usepackage{enumitem, kantlipsum}
\usepackage{xcolor,colortbl}
\usepackage{tikz}
\usepackage{multirow}
\usepackage{authblk}
\usepackage{collcell}
\usepackage[draft]{todonotes}
\usepackage{algorithm}
\usepackage{algorithmic}

\usepackage[pagebackref=true,breaklinks=true,letterpaper=true,colorlinks,bookmarks=false]{hyperref}

\cvprfinalcopy 

\def\cvprPaperID{1522} 

\ifcvprfinal\fi

\makeatletter
\renewcommand\AB@affilsepx{\qquad \protect\Affilfont}
\makeatother
\begin{document}

\title{Exploring Uncertainty in Conditional Multi-Modal Retrieval Systems}
\author[1]{Ahmed Taha}
\author[2]{Yi-Ting Chen}
\author[1]{Xitong Yang}
\author[2]{Teruhisa Misu}
\author[1]{Larry Davis}
\affil[1]{University of Maryland, College Park}
\affil[2]{Honda Research Institute, USA}
\affil[ ]{\tt\small \{ahmdtaha,xyang35,lsd\}@umiacs.umd.edu \qquad \tt\small \{ychen,tmisu\}@honda-ri.com}
\renewcommand\Authands{, and }


\maketitle

\begin{abstract}

We cast visual retrieval as a regression problem by posing triplet loss as a regression loss. This enables epistemic uncertainty estimation using dropout as a Bayesian approximation framework in retrieval. Accordingly, Monte Carlo (MC) sampling is leveraged to boost retrieval performance. Our approach is evaluated on two applications: person re-identification and autonomous car driving.  Comparable state-of-the-art results are achieved on multiple datasets for the former application. 

We leverage the Honda driving dataset (HDD) for autonomous car driving application. It provides multiple modalities and similarity notions for ego-motion action understanding. Hence, we present a multi-modal conditional retrieval network. It disentangles embeddings into separate representations to encode different similarities. This form of joint learning eliminates the need to train multiple independent networks without any performance degradation. Quantitative evaluation highlights our approach competence, achieving $6\%$ improvement in a highly uncertain environment.

\end{abstract}

\section{Introduction}

Quantifying uncertainty is vital when basing decisions on deep network outputs~\cite{ilg2018uncertainty,nair2018exploring}. Uncertainty can also improve network training and boost performance during inference. This has been demonstrated in both  regression~\cite{kendall2017uncertainties,feng2018leveraging,gal2016dropout} and classification~\cite{nair2018exploring,huang2018efficient,gal2016dropout} contexts. In this paper, we reformulate the triplet loss, a ranking loss, as a regression problem to explore uncertainty in a retrieval context. By enabling dropout during both training and inference, our formulation leverages Monte Carlo (MC) sampling for embedding. This boosts performance and enables embedding uncertainty estimation.


Gal and Ghahramani~\cite{gal2016dropout} utilized dropout as a Bayesian approximation for regression and classification applications. Building on top of that, we reformulate our retrieval ranking loss as a regression loss. While applied specifically on triplet loss, an extension to other ranking losses like quadruplet~\cite{huang2016local} and quintuplet~\cite{huang2016learning} is straightforward. We evaluate our formulation using two person re-identification datasets. State-of the art results are achieved on  DukeMTMC-reID and Market-1501 datasets. We further highlight our approach on an autonomous navigation application. This is a high uncertainty environment where large performance improvements are expected.


For autonomous navigation, we utilize the Honda driving dataset (HDD). It is designed to support learning driver actions. This dataset provides multiple modalities, \eg camera and CAN sensors. Driver actions have multiple similarities in terms of \textit{goal-oriented} and \textit{stimulus-driven} actions. Drivers plan goal-oriented actions like left and right turns to reach their destinations. In contrast, stimulus-driven actions are consequences of external factors like avoiding a parked car. A pair of actions, like a slow left-turn and a fast left-lane change, has multiple similarities. We introduce a conditional retrieval framework that learns an embedding per similarity. A conditional similarity network (CSN)~\cite{veit2017conditional} is leveraged to capture different similarities.

\begin{figure*}[ht!]
	\begin{center}
		\includegraphics[width=0.8\linewidth]{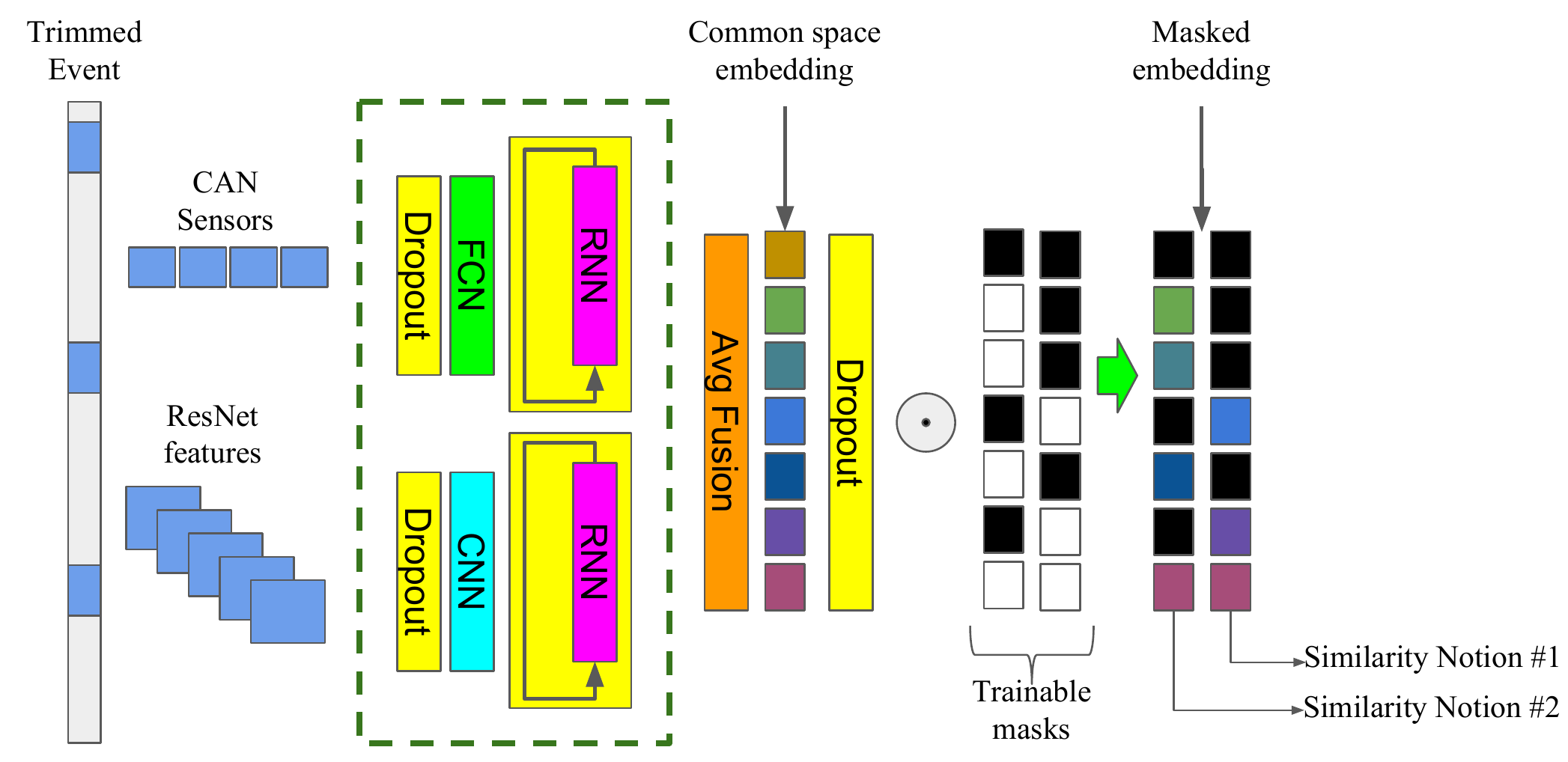}
	\end{center}
	
	\caption{Our proposed multi-modal conditional retrieval end-to-end network. Given a trimmed event, multiple samples are drawn. ResNet features, pre-extracted from video frames, and CAN sensors are independently embedded using separate encoders; then fused into a common space by averaging. Trainable masks enable conditional retrieval for the multiple similarity notions. A \textit{DropoutWrapper} is used for the RNN layers}
	\label{fig:arch}
\end{figure*}

 
Combining multi-modal data to boost system performance has drawn attention in various disciplines. In the medical field, patient heterogeneous scans are utilized to improve segmentation quality~\cite{chartsias2018multimodal,havaei2016hemis}. For autonomous navigation, multi-view~\cite{chen2017multi} and multi-sensor~\cite{cho2014multi} improve moving object detection and tracking respectively. Figure~\ref{fig:arch}  presents our multi-modal conditional retrieval framework. It employs average fusion to merge different modality embeddings. CSN supports multiple similarities
by disentangling the fused embedding into separate representations.



%



In summary, the key contributions of this paper are:
\begin{enumerate}[noitemsep]
	\item formulating triplet loss as a regression loss to enable MC sampling and uncertainty exploration in the retrieval context,
		\item an end-to-end multi-modal conditional retrieval system that promotes shared representation learning.
	\item leveraging Monte Carlo sampling during inference to boost retrieval performance. This achieves comparable state-of-the-art results on person re-identification datasets and significant improvement in autonomous car driving.
\end{enumerate}

\section{Related Work}

\subsection{Dropout as a Bayesian Approximation}
Given a supervised task with an input $X$ and ground-truth output $Y$, the Gaussian process (GP) ~\cite{rasmussen2004gaussian} allows modeling distributions over functions that generate the data. This enables model uncertainty estimation with Monte-Carlo (MC) sampling. To model the functions' distributions, a Bayesian approach is followed; starting with some prior distribution over the space of functions $p(f)$, we look for the posterior distribution over the space of functions
\begin{equation}
p(f|X,Y) \propto  P(Y|X,f) P(f) 
\end{equation}
$p(f|X,Y)$ evaluation requires choosing a covariance function $K(X1 , X2 )$. It defines the (scalar) similarity between every pair of input points $K(x_i,x_j)$. This involves an inversion of an $N \times N $ matrix, an $O(N^3)$ time complexity operation. Variational inference is an approximation approach with manageable time complexity. By conditioning the model on random variables $W$ instead of $f$, i.e. using $P(W|X,Y)$ instead of $P(f|X,Y)$, the predictive distribution for a new input point $x^\star $ is then given by
\begin{equation}
P({ y }^{ * }|{ x }^{ * },X,Y)=\int _{  }^{  }{ P({ y }^{ * }|{ x }^{ * },w)P(w|X,Y)~dw } 
\end{equation}
where $W$ acts as the weights of a neural network function. $P({ y }^{ * }|{ x }^{ * },X,Y)$ cannot be evaluated analytically, but an approximation using variational distribution $q(w)$ is possible. This leads to a  Kullback-Leibler $(KL)$ divergence minimization 
\begin{equation}
KL(q(w)|p(w|X,Y))
\end{equation}
Minimizing the Kullback-Leibler divergence is equivalent to maximizing the log evidence lower bound (ELBO) ~\cite{bishop2006pattern}
\begin{equation}
\L_{vi} =\int _{  }^{  }{ q(w)\log { P(Y|X,w)dw } -KL(q(w)||p(w)) } 
\end{equation}
Gal and Ghahramani~\cite{gal2016dropout} show that the ELBO objective function is equivalent to optimizing a regression neural network
\begin{equation}
\L_{reg} =-\frac { 1 }{ 2N } \sum _{ n=1 }^{ N }{ { || { y }_{ n }- { \hat { y }  }_{ n }|| }_{ 2 }^{ 2 } } -\lambda  { ||M|| }_{ 2 }^{ 2 }
\end{equation}
where $\lambda$ is a weight regularization hyper-parameter, the random variable realization $W$ can be approximated as $W = zM$ where $z \sim  Bernoulli(P)$, i.e. dropout is required before every weight layer. Flipping the $-ve$ sign for $\L_{reg}$ leads to the standard minimization objectives.

This enables MC sampling from dropout networks outputs to approximate the posterior $P(Y| X,W)$ for any regression function. In the next section, we cast triplet loss as a regression loss to allow MC sampling in the retrieval context.

\subsection{Conditional Retrieval System}
Most multi-modal fusion methods produce a single embedding space. This promotes robustness to missing modalities ~\cite{havaei2016hemis}. Yet, a single space embedding poses limitations for capturing multiple similarities. Veit \etal~\cite{veit2017conditional} propose a conditional similarity network (CSN) that compromises between the single embedding space and multiple similarity support. CSN learns a single embedding space, and then disentangles a per-similarity representation. 

CSN eliminates the requirement to train individual specialized networks for each similarity while promoting shared representation learning. Based on their work, we learn embedding masks to model different similarities. This approach reduces system complexity by training a single network. It also boosts retrieval performance due to the joint formulation and shared representation learning.

\subsection{Multi-Modal Fusion}

For a multi-modality system with $K$ modalities $(M_1,M_2,...,M_k)$, multiple fusion approaches exist.
Element-wise addition is the most trivial approach. Similar to addition,~\cite{havaei2016hemis} concatenates the first and second moments, mean and variance, across modalities. This fusion variant is robust to any combinatorial subset of available modalities provided as input. This eliminates the need to learn a combinatorial number of imputation models. Other fusion methods, including gated~\cite{arevalo2017gated}, max ~\cite{chartsias2018multimodal} and bilinear~\cite{fukui2016multimodal} fusion, are employed in integrating natural language and visual modalities. Similar to ~\cite{havaei2016hemis}, we leverage mean fusion for its simplicity and robustness.

\section{Bayesian Retrieval}
This section describes our approach. We first discuss how triplet loss can be cast as a regression loss. Then, we discuss how epistemic uncertainty is employed in retrieval context. Finally, we present the architectural designs for the two applications. 

\subsection{Triplet Loss as Regression Loss}

Triplet loss~\cite{schroff2015facenet} is a similarity embedding metric.  It is more efficient compared to contrastive loss~\cite{li2017improving}, and less computationally expensive than quadruplet~\cite{huang2016local,chen2017beyond} and quintuplet~\cite{huang2016learning} losses. It has been applied successfully for face recognition~\cite{schroff2015facenet,sankaranarayanan2016triplet} and person re-identification~\cite{cheng2016person,su2016deep,ristani2018features}. In this section, we cast this ranking loss as a regression function to exploit uncertainty estimation in retrieval.

In~\cite{gal2016dropout}, Gal and Ghahramani propose an approach to uncertainty estimation by training a dropout network. Model uncertainty is estimated by computing the first two moments of the predictive distribution -- output mean and variance. Training with dropout had previously been approached using model averaging~\cite{srivastava2014dropout} by taking the prediction first moment. It scales the weights proportionally to the dropout percentage during training and removes dropout during inference.


Gal and Ghahramani~\cite{gal2015dropout} show that deep NNs with dropout applied before every weight layer are mathematically equivalent to approximate variational inference of a deep Gaussian process. Their derivation applies to networks trained with either the Euclidean loss for regression or softmax loss for classification. We build on their work by casting triplet loss as a trivariate regression function. This enables MC sampling to model uncertainty in retrieval.

Given a training dataset containing $N$ triplets $\{(x_1,y_1,z_1),(x_2,y_2,z_2),...(x_n,y_n,z_n)\}$ and their corresponding outputs $(d_1,...,d_n)$, the triplet loss can be formulated as a trivariate regression function as follows
\begin{align}\label{eq:tri_func}
& f_{tri}(x_i,y_i,z_i)= d_i \in [0,2+m]\\
&={ \left[D(\left\lfloor x_i \right\rfloor ,\left\lfloor y_i \right\rfloor )-D(\left\lfloor x_i \right\rfloor,\left\lfloor z_i \right\rfloor) +m \right]  }_{ + } \label{eq:tri_func_full}
 \end{align}
where ${ \left[ . \right]  }_{ + }= max(0,.)$, $m$ is the margin between different classes embedding. $\left\lfloor. \right\rfloor$ and $D(,)$ are the unit-length embedding and the Euclidean distance functions respectively. $f_{tri}(x_i,y_i,z_i)$ outputs $d_i=0$ if $y_i,x_i \in c_i$ and $z_i \in c_j$; and $d_i=2+m$ if $z_i,x_i \in c_i$ and $y_i \in c_j$  s.t. $i \ne j$. 

Equations~\ref{eq:k_tup_func} and ~\ref{eq:k_tup_func_full}  provide a generalization for k-tuplets where $k \ge 3$
\begin{align}\label{eq:k_tup_func}
& f_{k\_tup}(x_0,..,x_j,..,x_k)= d \in [0,2+(k-2)\times m]\\
&=\sum _{ j=0 }^{ k-2 }{ \left[ D(\left\lfloor x_{ 0 } \right\rfloor ,\left\lfloor x_{ j+1 } \right\rfloor )-D(\left\lfloor x_{ 0} \right\rfloor ,\left\lfloor x_{ j+2 } \right\rfloor )+m \right]  }_{ + }  \label{eq:k_tup_func_full}\\
& \text{s.t.} \quad D(\left\lfloor x_{ 0 } \right\rfloor ,\left\lfloor x_{ 1 } \right\rfloor ) < D(\left\lfloor x_{ 0 } \right\rfloor ,\left\lfloor x_{ j } \right\rfloor ) <  D(\left\lfloor x_{ 0 } \right\rfloor ,\left\lfloor x_{ k } \right\rfloor )
\end{align}

In our experiments, the triplet loss is adopted for its lower complexity. We evaluate it with two sampling strategies: hard-mining for person re-identification~\cite{hermans2017defense} and semi-negative sampling~\cite{schroff2015facenet} for autonomous car driving.

\subsection{Epistemic Uncertainty as Data-Independent Uncertainty}
Kendall and Gal~\cite{kendall2017uncertainties} identify two main types of network uncertainty: epistemic and aleatoric. Epistemic captures network uncertainty in terms of generalization -- what the training data omit; aleatoric captures uncertainty regarding information which the training data cannot explain -- why the network is underfitting. Gal and Ghahramani~\cite{gal2016dropout} leverage epistemic uncertainty for model selection, i.e. hyper-parameter tuning. Kendall \etal~\cite{kendall2017multi} leverage aleatoric uncertainty  to improve multi-task learning by estimating pseudo optimal weights for each task's loss term. In this work, we leverage epistemic uncertainty to improve retrieval performance.


Using our triplet loss formulation, MC sampling is employed to estimate the embedding uncertainty. First, the model is trained with a dropout before every weight layer.  During inference, dropout is left enabled to sample from the approximate posterior~\cite{kendall2017uncertainties} --  stochastic forward passes, referred to as Monte Carlo dropout. Multiple passes, through our dropout enabled network, generate multiple embeddings per sample $x_i$. These embeddings are aggregated using the first moment into the final embedding for retrieval, $emb(x_{ i })=\frac{1}{MC} \sum _{ i=1 }^{ MC }{ \left\lfloor { x }_{ i } \right\rfloor  } $. The second moment indicates the model uncertainty.

\subsection{Network Architecture}
Two different architectures are employed to evaluate our approach. For person re-identification, Hermans' \etal~\cite{hermans2017defense} single network architecture with triplet loss is utilized with a few modifications. Resnet-50~\cite{he2016deep} is replaced by DenseNet-169~\cite{huang2017densely} for its built-in dropout layers~\cite{kendall2017uncertainties}. A dropout layer is employed after convolutional layers~\cite{kendall2017uncertainties,gal2015bayesian} and before fully connected layers~\cite{gal2016dropout}. By default, the network embeddings are normalized to the unit-circle. 

For autonomous navigation, we propose a multi-modal conditional retrieval network depicted in figure~\ref{fig:arch}. The architecture strives for efficiency and simplicity with a single loss term -- the triplet loss. Each modality is independently encoded, then fused into a common space using mean fusion. A set of trainable masks enable conditional retrieval for multiple similarities. As illustrated in our experiments, this form of joint learning improves efficiency by leveraging shared representation learning~\cite{caruana1997multitask,kaiser2017one,kokkinos2017ubernet,luo2018fast}.

Modeling event temporal context provides an additional and important clue for action understanding~\cite{simonyan2014two}. For temporal context encoding, different variants exists~\cite{simonyan2014two,dai2017temporal,yu2015fast,taha2018two}. In our work, a recurrent neural network (RNN)~\cite{funahashi1993approximation,hochreiter1997long} with a \textit{DropoutWrapper} is employed within each encoder. During training, three random samples are drawn from an event. They are independently encoded then temporally fused using RNNs. More samples per event improve performance. Unfortunately, GPU memory constrains the number of samples. In the case of Honda driving dataset (HDD), random samples are frames and sensor measurements from the camera and CAN sensor streams respectively. To reduce GPU memory requirements, a per-frame representation is extracted from the Conv2d\_7b\_1x1 layer of InceptionResnet-V2~\cite{szegedy2017inception} pretrained on ImageNet. Further architectural details are described in the supplementary material.
\section{Experiments}


Evaluation is performed using an image retrieval applications (person re-identification) and a video retrieval application (autonomous navigation ``first-person" videos). By default, the triplet loss margin $m=0.2$~\cite{schroff2015facenet}. Similar to~\cite{kendall2017uncertainties}, 50 Monte Carlo samples are used during inference. In our experiments, a baseline refers to training with and inferring without dropout~\cite{srivastava2014dropout}. All experiments are conducted on Titan Xp 12 GB GPU.

\subsection{Person Re-Identification}
Person re-identification is employed in Multi-Target Multi-Camera Tracking system. It provides a quantitative benchmark to evaluate our triplet loss casting. An ID system retrieves images of people and ranks them by decreasing similarity to a given person query image. Two datasets are used for evaluation: DukeMTMC-reID~\cite{ristani2016performance,zheng2017unlabeled} and Market-1501~\cite{zheng2015scalable}. \textbf{DukeMTMC-reID} includes 1,404 identities appearing in more than two cameras and 408 identities appearing in a single camera for distraction purpose. 702 identities are reserved for training and 702 for testing. \textbf{Market-1501} is a large-scale person re-identification dataset with 1,501 identities observed by 6 near- synchronized cameras.

For each training mini-batch, we uniformly sample $P = 18$ person identities without replacement.  For each person, $K = 4$ sample images are drawn without replacement with resolution $256\times 128$. The learning rate is $3*10^{-4}$ for the first 15000 iterations, and decays to $10^{-7}$ at iteration 25000. A dropout rate $p=0.15$ is employed. Ristani and Tomasi~\cite{ristani2018features} proposed a set of person re-identification specific augmentation techniques to boost performance, \eg hide small rectangular image patches to simulate occlusion. Since our main objective is exploring uncertainty within the retrieval context, all approaches are evaluated without augmentation during training or testing.

To highlight our approach generality, both vanilla triplet loss and adaptive weight triplet loss (AWTL)~\cite{ristani2018features} are employed for evaluation. AWTL improves the hard-sampling strategy by assigning small and large weights to easy and difficult triplets respectively. We adapt ~\cite{ristani2018features} configurations in AWTL evaluation -- unnormalized embedding with soft-margin.

We follow Hermans \etal~\cite{hermans2017defense} evaluation procedure with a few architectural modifications. Similar to~\cite{kendall2017uncertainties}, we use DenseNet-169~\cite{huang2017densely} for its dropout support. Dropout layers are added after convolutional~\cite{gal2015bayesian} and before fully connected layers. Tables~\ref{tbl:market} and~\ref{tbl:duke} present our quantitative evaluation where comparable state-of-the-art results are achieved.  The two key reasons for this improvement are (1) using a strong neural network, DenseNet, and (2) leveraging MC sampling during inference to raise performance even further.



\begin{table}[]
	\centering
	\begin{tabular}{|l|c|c|}
		\hline
	Method          & mAp &  Top 1  \\ \hline
	PointSet~\cite{zhou2017point}    &   44.27 &   70.72	 \\ \hline
	SomaNet~\cite{barbosa2018looking}    &   47.89 &   73.87	 \\ \hline
	PAN~\cite{zheng2018pedestrian}    &   63.35 &   82.81	 \\ \hline \hline


	Tri-ResNet~\cite{hermans2017defense}    &   66.63 &   82.99	 \\ \hline 
	Tri-Dense  (baseline) &   68.62 &   83.94	 \\ \hline 
    \textbf{Tri-Dense + 50 MC (ours)} &  \textbf{68.81} &   \textbf{84.03}	 \\ \hline \hline
    
    Tri-ResNet + AWTL~\cite{ristani2018features}   &   68.03 &   84.20	 \\ \hline 
    Tri-Dense + AWTL   &   71.45&   84.89	 \\ \hline
    \textbf{Tri-Dense  + AWTL  +  50 MC (ours)}  &   \textbf{72.19} &   \textbf{85.87}	 \\ \hline

	\end{tabular}
	
	\caption{Quantitative evaluation on Market-1501. MC indicates the number of Monte Carlo samples drawn during inference. Dense indicates a DenseNet with dropout layers. }
	\label{tbl:market}
\end{table}

\begin{table}[]
	\centering
	\begin{tabular}{|l|c|c|}
		\hline
		Method          & mAp &  Top 1  \\ \hline
		Baseline~\cite{zheng2016person}    &   44.99 &   65.22	 \\ \hline
		PAN~\cite{zheng2018pedestrian}    &   51.51 &   71.59	 \\ \hline
		SVDNet~\cite{sun2017svdnet}    &   56.80 &   76.70	 \\ \hline \hline

		
		Tri-ResNet~\cite{hermans2017defense}    &   54.60 &   73.24	 \\ \hline 
		Tri-Dense  (baseline) &   59.97 &   76.97	 \\ \hline 
		\textbf{Tri-Dense + 50 MC (ours)} &   \textbf{61.01} &  \textbf{78.10}	 \\ \hline \hline
		
		Tri-ResNet + AWTL~\cite{ristani2018features}   &   54.97 &   74.23	 \\ \hline 
		Tri-Dense +  AWTL &   61.89 &   79.35	 \\ \hline 
		\textbf{Tri-Dense +  AWTL +  50 MC  (ours)}  &   \textbf{62.88} & \textbf{79.76}	 \\ \hline 
	\end{tabular}
	
	\caption{Quantitative evaluation on DukeMTMC-ReID.}
	\label{tbl:duke}
\end{table}

\subsection{Autonomous Car Driving}

\begin{figure}
	\begin{tikzpicture} \begin{axis}[ ybar,width=0.5\textwidth, height=3.0cm, enlargelimits=0.15, 
	symbolic x coords={
		Stop 4 sign,
		Stop 4 light,
		Stop 4 congestion,
		Stop for others,
		Stop 4 pedestrian,
		Avoid parked car
	}, xtick=data, nodes near coords align={vertical}, x tick label style={rotate=45,anchor=east},] 
	\addplot coordinates {
		(Stop 4 sign,2322)
		(Stop 4 light,754) 
		(Stop 4 congestion,1943)
		(Stop for others,50)
		(Stop 4 pedestrian,100)
		(Avoid parked car,140) 
	};  \end{axis}
	\end{tikzpicture}
	\caption{HDD imbalance stimulus-driven actions distribution.}
	\label{fig:stimuli_dist}
\end{figure}
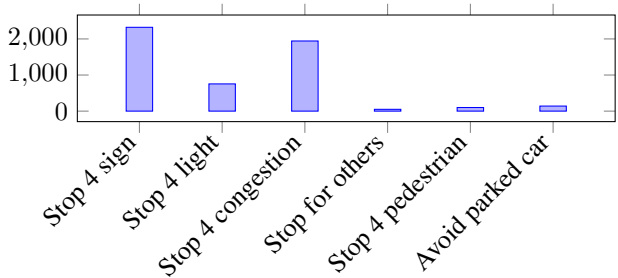

The Honda driving dataset (HDD)~\cite{ramanishka2018toward} is designed to support modeling driver behavior and causal reasoning understanding. It defines four annotation layers: goal-orientated actions, stimulus-driven actions, cause, and attention. \textbf{Goal-oriented actions} represents the egocentric activities taken to reach a destination like left and right turns. \textbf{Stimulus-driven} are actions due to external causation factors like stopping to avoid a pedestrian or stopping for a traffic light. \textbf{Cause} indicates the reason for an action. Finally, the \textbf{attention} layer localize the traffic participants that drivers attend to. Every layer is categorized into a set of classes (actions).  Figures~\ref{fig:HDD_dist} and~\ref{fig:stimuli_dist} show the class distribution for  goal and stimulus layers respectively. 

HDD is a multimodal dataset with camera video, LiDAR, GPS, IMU and CAN sensor signals. In our experiment, we use two modalities: the camera video, and the CAN sensor signals. The CAN sensor records the throttle angle, brake pressure, steering angle, yaw rate and speed at 100 Hz. These measurements are powerful for understanding goal-orientated actions, \eg \textit{left} and \textit{right turns}. Yet, they are very limited in terms of visual and external factors comprehension-- pedestrian or traffic behavior.

To understand the experiment results, we elaborate further on the dataset details. Both camera and CAN modalities are capable of learning most goal-oriented actions like \textit{left} and \textit{U-turns}. Thus, a neural network can easily learn goal-oriented actions and retrieve events with low uncertainty. In contrast, a network trained with CAN modality to learn stimulus-driven actions will suffer high uncertainty due to its visual limitations. Further dataset details are described in the supplementary material. 

 For autonomous navigation experiments, the learning rate is $lr=0.01$ for the first 250 epochs, and decays linearly to zero at epoch 500. Adam optimizer~\cite{kingma2014adam} is utilized.  $p=0.1$ and $d=128$ are the dropout rate and embedding space dimensionality. To evaluate our approach, all test split events are embedded into the unit-circle. An event retrieval evaluation using query-by-example is performed. Given a query event, similarity scores to all events are computed --i.e. a leave-one-out cross evaluation on the test split. Performances of all queries are averaged to obtain the final evaluation. 

To tackle data imbalance and highlight performance on minority classes, both micro and macro average accuracies are reported. Macro-average computes the metric for each class independently before taking the average. Micro-average is the traditional mean for all samples. Macro-average treats all classes equally while micro-averaging favors majority classes. Tables~\ref{tbl:goal_lbl},~\ref{tbl:stimuli_lbl}, and~\ref{tbl:both_lbl} show quantitative evaluation for networks trained on goal, stimulus and both goal and stimulus respectively. The baseline measures retrieval performance when dropout is enabled during training only~\cite{srivastava2014dropout}. Our MC sampling approach measures the performance when dropout is enabled during both training and inference. MC indicates the number of Monte Carlo samples utilized to compute the final embedding.

The quantitative evaluation shows that our approach always outperforms the baseline. This highlights the utility of epistemic uncertainty in retrieval. A larger performance margin ($\triangle> 6\%$) is achieved when retrieving stimulus-driven actions. This aligns with recent literature findings where epistemic utility is marginal for low uncertainty tasks~\cite{ilg2018uncertainty} while significant for high uncertainty tasks like LiDAR 3D Object Detection~\cite{feng2018leveraging}. Table~\ref{tbl:both_lbl} demonstrates how jointly learning both similarities outperforms specialized networks in tables~\ref{tbl:goal_lbl} and~\ref{tbl:stimuli_lbl}. Our conditional retrieval network benefits from learning all concepts jointly within one model, utilizing the shared structure between concepts~\cite{veit2017conditional}.

It is straightforward to use Monte Carlo sampling to measure a particular embedding uncertainty using the second moment. This is employed in semantic segmentation and optical flow estimation~\cite{kendall2017multi,ilg2018uncertainty}. In retrieval, a more appealing measure is the system uncertainty per-class. Table~\ref{tbl:epistemic_analysis} presents per-class uncertainty normalized by class size to handle data imbalance. A high correlation between system uncertainty and class size is the first observation. CAN sensors, like steering angle, play a key role in reducing \textit{left}, \textit{right} and \textit{U-turns} uncertainty. Intersection passing is challenging for CAN sensors. Yet, it is a majority classes and this explains its mid-range variance. 

Goal-oriented and stimulus-driven layers have ten and six actions respectively. Thus, stimulus actions uncertainty is relatively small compared to goal-oriented. Yet, the high uncertainty pattern manifests itself  for minority classes. \textit{Stop for sign} has the lowest uncertainty because it is a majority class and it can be inferred from speed sensor measurements. Despite being majority classes, \textit{stop for light} and \textit{stop for congestions} suffer higher uncertainty compared to \textit{stop for sign}. CAN sensors limitations differentiating these two classes explains their relative high variance.


\begin{figure}
	\begin{tikzpicture} \begin{axis}[ ybar,width=0.5\textwidth, height=3.0cm, enlargelimits=0.15, symbolic x coords={Intersection Pass,Left Turn,Right Turn,Left Lane Change,Right Lane Change,Cross-walk passing,U-Turn,Left Lane Branch,Right Lane Branch,Merge}, xtick=data, nodes near coords align={vertical}, x tick label style={rotate=45,anchor=east},] 
	\addplot coordinates {
		(Intersection Pass,5651)
		 (Left Turn,1689) 
		 (Right Turn,1677)
		  (Left Lane Change,560)
		   (Right Lane Change,518)
		 (Cross-walk passing,182) 
		 (U-Turn,85)
		  (Left Lane Branch,235) 
		  (Right Lane Branch,93) 
		  (Merge,143)
	};  \end{axis}
	\end{tikzpicture}
	\caption{HDD long tail goal-oriented actions distribution.}
	\label{fig:HDD_dist}
\end{figure}
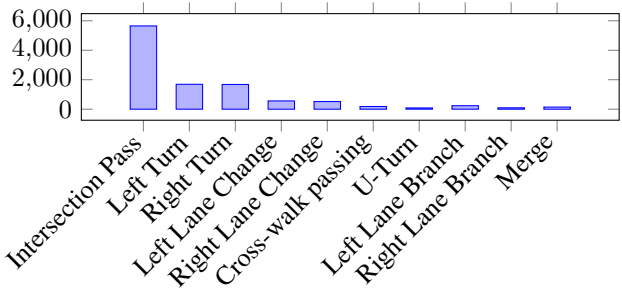

\begin{table}[]
	\centering
	\begin{tabular}{|l|c|c|}
	\hline
	Method          & Micro mAp &  Macro mAp \\ \hline
	Baseline    &   86.15 &   48.20	 \\ \hline
	Epistemic (MC=25)      &   86.25 &  48.62 \\ \hline
	Epistemic (MC=50)         &    \textbf{86.38}  & \textbf{48.68}\\ \hline
\end{tabular}

\caption{Goal-based retrieval quantitative evaluation using both camera and CAN sensors modalities.}
	\label{tbl:goal_lbl}
\end{table}

\begin{table}[]
	\centering
	\begin{tabular}{|l|c|c|}
		\hline
		Method          & Micro mAp &  Macro mAp \\ \hline
		Baseline    &   70.62 &   40.61	 \\ \hline
		Epistemic (MC=25)      &  76.78  &  44.97 \\ \hline
		Epistemic (MC=50)         &  \textbf{76.88}  & \textbf{45.03} \\ \hline
	\end{tabular}
	
	\caption{Stimulus-based retrieval quantitative evaluation using both camera and CAN sensors modalities.}
	\label{tbl:stimuli_lbl}
\end{table}

\begin{table}[]
	\centering
	\begin{tabular}{|c|l|c|c|}
		\hline
		& Method          & Micro mAp &  Macro mAp \\ \hline
		{\multirow{3}{*}{\rotatebox[origin=c]{90}{Goal}}} & Baseline    &   86.11 &   48.67	 \\ \cline{2-4}
		&Epistemic (MC=25)     &   86.09 &  48.66 \\ \cline{2-4}
		&Epistemic (MC=50)         &   \textbf{86.22}  & \textbf{49.14} \\ \hline \hline
		
		{\multirow{3}{*}{\rotatebox[origin=c]{90}{Stimulus}}} &Baseline    &   71.98 &   41.24	 \\ \cline{2-4}
		&Epistemic (MC=25)     &   \textbf{78.00} &  44.92 \\ \cline{2-4}
		&Epistemic (MC=50)         &   77.94   & \textbf{44.96} \\ \hline
	\end{tabular}

	\caption{Joint based retrieval, both goal and stimulus, quantitative evaluation using both camera and CAN sensors modalities.}
		\label{tbl:both_lbl}
\end{table}

\newcommand*{\MinNumber}{0.0}%
\newcommand*{\MaxNumber}{35}%

\newcommand{\ApplyGradient}[1]{%
	\pgfmathsetmacro{\PercentColor}{max(min(100.0*(#1 -\MinNumber)/(\MaxNumber-\MinNumber),100.0),0.00)} 
	\edef\x{\noexpand\cellcolor{blue!\PercentColor}}\x\textcolor{white}{#1}}

\newcommand*{\MinNumberII}{0.0}%
\newcommand*{\MaxNumberII}{25}%

\newcommand{\ApplyGradientII}[1]{%
	\pgfmathsetmacro{\PercentColor}{max(min(100.0*(#1 -\MinNumberII)/(\MaxNumberII-\MinNumberII),100.0),0.00)} 
	\edef\x{\noexpand\cellcolor{red!\PercentColor}}\x\textcolor{white}{#1}}

\begin{table}[ht]
	\begin{center}
		\begin{tabular}{|l|c|l|c|}
			\hline
			Goal & Var  & Stimulus &  Var\\ \hline
			Intersection Pass & \ApplyGradient{17.80} & Stop 4 Sign & \ApplyGradientII{12.74}  \\ \hline
			Left (L) Turn & \ApplyGradient{11.71} & Stop 4 Light &  \ApplyGradientII{17.59} \\ \hline
			Right (R) Turn & \ApplyGradient{12.69} & Stop 4 Congestion & \ApplyGradientII{15.52} \\ \hline
			L LN Change & \ApplyGradient{26.28} & Stop for others  &  \ApplyGradientII{21.42} \\ \hline
			R LN Change & \ApplyGradient{24.35} & Stop 4 Pedestrian  &   \ApplyGradientII{21.67} \\ \hline
			Crosswalk Pass& \ApplyGradient{26.80} & Avoid Parked Car & \ApplyGradientII{23.61}  \\ \hline
			U-Turn & \ApplyGradient{18.21} &  &  \\ \cline{0-1}
			L LN Branch & \ApplyGradient{28.14} &  &  \\ \cline{0-1}
			R LN Branch & \ApplyGradient{31.71} &  &  \\ \cline{0-1}
			Merge & \ApplyGradient{30.62} &  &  \\ \hline
		\end{tabular}
	\end{center}
\caption{Uncertainty evaluation per class. Variance in the second and fourth columns are multiplied by $10^{3}$.}
	\label{tbl:epistemic_analysis}
\end{table}

\subsection{Ablation Study}\label{sec:ablation}
While both tables~\ref{tbl:goal_lbl} and~\ref{tbl:stimuli_lbl} show improvement using Monte Carlo sampling, the margin is larger for the stimulus similarity. CAN sensors, like brake pressure and speed, have clear limitation differentiating stimulus events like \textit{stop for congestion} versus \textit{stop for a traffic light}. In this subsection, we validate this hypothesis by removing the CAN modality and quantitatively re-evaluating our system for both goal and stimulus similarity labels.

Tables~\ref{tbl:res_goal_lbl} and~\ref{tbl:res_stimuli_lbl} illustrate the effect of dropping the CAN modality. While Monte Carlo sampling improvement persists, the margins are comparable for both goal-oriented and stimulus-driven actions. Removing CAN reduces system uncertainty, thus the Monte Carlo sampling benefits drop. This is consistent with recent results, where small and large improvements are achieved in low~\cite{ilg2018uncertainty} and high~\cite{feng2018leveraging} uncertainty tasks respectively. Far and occluded 3D objects detection using LiDAR sensors~\cite{feng2018leveraging} benefits significantly from uncertainty evaluation. Exploiting epistemic for high uncertainty tasks achieves larger improvements compared to low uncertainty tasks.

  
The second finding is that dropping CAN has a bigger impact on goal versus stimulus retrieval. CAN sensors like steering angle and throttle angle are informative for goal-oriented actions understanding. We attribute the stimulus performance drop to a latent correlation between CAN sensors and some stimulus-driven actions. For example, multiple speed sensor measurements across an event can reveal driver's different behaviors at stop-signs versus congestions. The speed drops momentarily at a stop sign but lasts longer for congestion and traffic lights.

\begin{table}[]
	\centering
	\begin{tabular}{|l|c|c|}
		\hline
		Method          & Micro mAp &  Macro mAp \\ \hline
		Baseline    &   80.96 &   42.05	 \\ \hline
		Epistemic (MC=50)         &    \textbf{81.20}  & \textbf{42.33}\\ \hline
	\end{tabular}
	
	\caption{Goal-based retrieval quantitative evaluation using the camera modality.}
	\label{tbl:res_goal_lbl}
\end{table}

\begin{table}[]
	\centering
	\begin{tabular}{|l|c|c|}
		\hline
		Method          & Micro mAp &  Macro mAp \\ \hline
		Baseline    &   \textbf{66.63} &   37.53	 \\ \hline
		Epistemic (MC=50)         &  \textbf{66.63}  & \textbf{37.69} \\ \hline
	\end{tabular}
	
	\caption{Stimulus-based retrieval quantitative evaluation using the camera modality.}
	\label{tbl:res_stimuli_lbl}
\end{table}

\subsection{Comparing Weight Averaging and Monte Carlo	Dropout Sampling}
\begin{figure}
	\centering	
	\pgfplotsset{width=\columnwidth,height=3.5cm}
	\begin{tikzpicture}
	\begin{axis}
	[ xlabel={
		No. of MC samples},
	ylabel={Goal mAp},
	legend entries={MC Sampling,Baseline},
	legend style={at={(1.0,0.7)}}
	] 
	
	
	\addplot[dashed,thick,blue] coordinates {  
		(1,84.46)  
		(2,85.34)
		(3,85.83) 
		(4,85.87) 
		(5,86.00)
		(6,86.11)
		(7,86.14)
		(8,86.11)
		(9,86.13)
		(10,86.13)
		(15,86.31)
		(20,86.29)
		(25,86.29)
		(30,86.37)
		(40,86.43)
		(50,86.36)
		(100,86.38)
	}; 
	
	\addplot[solid,thick,red] coordinates {  
		(1,86.15)  
		(2,86.15)
		(3,86.15) 
		(4,86.15) 
		(5,86.15)
		(6,86.15)
		(7,86.15)
		(8,86.15)
		(9,86.15)
		(10,86.15)
		(15,86.15)
		(20,86.15)
		(25,86.15)
		(30,86.15)
		(40,86.15)
		(50,86.15)
		(100,86.15)
	}; 
	\end{axis}
	\end{tikzpicture}
	\begin{tikzpicture}
	\begin{axis}
	[ xlabel={
		No. of MC samples},
	ylabel={Stimulus mAp},
	legend entries={MC Sampling,Baseline},
	legend style={at={(1.0,0.7)}}
	]

	
	\addplot[dashed,thick,blue] coordinates {  
		(1,74.14)  
		(2,75.82)
		(3,75.91) 
		(4,76.36) 
		(5,76.34)
		(6,76.35)
		(7,76.45)
		(8,76.48)
		(9,76.50)
		(10,76.45)
		(15,76.55)
		(20,76.84)
		(25,76.79)
		(30,76.80)
		(40,76.96)
		(50,76.88)
		(100,76.97)
	}; 
	
	\addplot[solid,thick,red] coordinates {  
		(1,70.62)  
		(2,70.62)
		(3,70.62) 
		(4,70.62) 
		(5,70.62)
		(6,70.62)
		(7,70.62)
		(8,70.62)
		(9,70.62)
		(10,70.62)
		(15,70.62)
		(20,70.62)
		(25,70.62)
		(30,70.62)
		(40,70.62)
		(50,70.62)
		(100,70.62)
	}; 


	
	\end{axis} 
	\end{tikzpicture}
	\caption{Monte Carlo sampling evaluation compared to weight averaging on two networks. The top and bottom networks are independently trained for goal and stimulus retrieval respectively. Both networks use the camera and CAN sensor modalities as input.}
	\label{fig:time_analysis}
\end{figure}
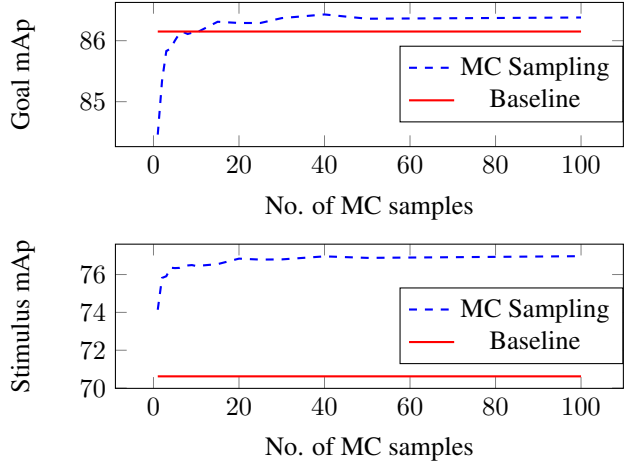

Monte Carlo sampling qualitatively allows us to understand model uncertainty and quantitatively boosts  performance. In this section, we compare the MC sampling approach against the weight averaging technique~\cite{srivastava2014dropout} -- our baseline. Figure~\ref{fig:time_analysis} depicts two systems for goal and stimulus retrieval, both trained with  the camera and CAN sensor modalities as input. MC sampling outperforms weight averaging in goal-based retrieval as the number of samples increases. This finding is consistent with Kendall \etal~\cite{kendall2015bayesian} where the baseline is outperformed  after nearly 10 samples and performance stabilizes beyond 50 samples. 

The stimulus retrieval performance is more interesting. MC sampling outperforms the baseline using a single sample when dropout is enabled. Table~\ref{tbl:details_stimulus} presents a detailed performance analysis to explain this phenomena. As the number of samples increases, a reasonable improvement is achieved for \textit{stop for sign} and minority actions. Yet, a larger margin is attained for both \textit{stop for light} and \textit{stop for congestion} actions. Both actions are majorities suffering high uncertainty from the CAN sensor perspective. As previously mentioned, a latent correlation can help detect \textit{stop for signs}, \eg momentarily speed drop. Yet, no similar pattern can disentangle \textit{stop for light} from \textit{stop for congestion} without visual cues. 


MC sampling cancels noise out through aggregating multiple high uncertain predictions. It is worth noting that this performance improvement comes with a higher computational cost during inference, i.e. multiple feed-forward passes are required. For an offline retrieval system, this can be acceptable, especially with parallelization workarounds. For a real-time system, a trade-off between performance and the number of samples is inevitable.

\newcommand*{\StimuliMinNumber}{1.0}%
\newcommand*{\StimuliMaxNumber}{10}%

\newcommand{\StimuliApplyGradient}[1]{%
	\pgfmathsetmacro{\PercentColor}{max(min(100.0*(#1 -\MinNumber)/(\MaxNumber-\MinNumber),100.0),0.00)} 
	\edef\x{\noexpand\cellcolor{blue!\PercentColor}}\x\textcolor{white}{#1}}

\begin{table}[h!]
	\centering
	\begin{tabular}{|l|c|c|c|}
		\hline
		Method            & Baseline & MC=1 & MC=100  \\ \hline
		Micro mAP               &     70.62     &   74.11   & \textbf{76.95}      \\ \hline
		Macro mAP         &     40.61     &   42.29   &     \textbf{45.13}  \\ \hline \hline
		
		\cellcolor{yellow!20}Stop  4 Sign       &\cellcolor{yellow!20}       85.59   &\cellcolor{yellow!20}    85.60  &     \cellcolor{yellow!20} \textbf{87.42}   \\ \hline
		
		\cellcolor{yellow!100}Stop 4 Light      & 		\cellcolor{yellow!100}      59.22   & 		\cellcolor{yellow!100}   66.09  &   		\cellcolor{yellow!100}  \textbf{70.11}   \\ \hline
		
		\cellcolor{yellow!80}Stop 4 Congestion &    \cellcolor{yellow!80} 71.63     &  \cellcolor{yellow!80} 77.15    &   \cellcolor{yellow!80} \textbf{80.35} \\ \hline
		
		\cellcolor{yellow!30}Stop 4 Others     &  \cellcolor{yellow!30}    2.37    &   \cellcolor{yellow!30}  3.65  &  \cellcolor{yellow!30}   \textbf{5.42}  \\ \hline
		
		\cellcolor{yellow!20}Stop 4 Pedestrian &   \cellcolor{yellow!20}   4.53    &  \cellcolor{yellow!20}  4.75  &   \cellcolor{yellow!20}   \textbf{5.87} \\ \hline
		
		\cellcolor{yellow!10}Avoid Parked Car  &       \cellcolor{yellow!10}20.34   &    \cellcolor{yellow!10}16.48  &    \cellcolor{yellow!10}\textbf{21.63}    \\ \hline
		
	\end{tabular}
	\caption{Detailed performance analysis on stimulus-driven actions. Row color emphasizes the  improvement margin. }
	\label{tbl:details_stimulus}
\end{table}

\begin{figure*}[h!]
	\begin{subfigure}{1.0\textwidth}
		\centering
		\setlength{\fboxsep}{0pt}%
		\setlength{\fboxrule}{2pt}%
		
		\fcolorbox{blue}{white}{\includegraphics[width=.19\linewidth]{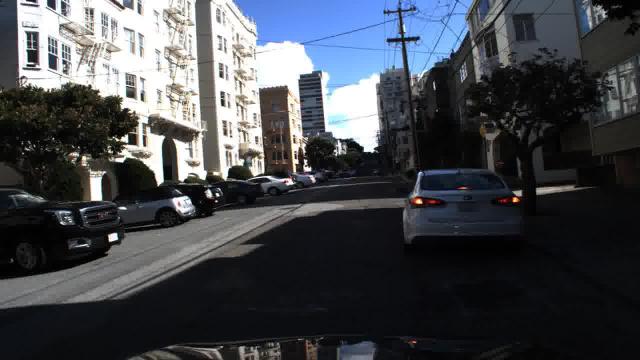}
			\includegraphics[width=.19\linewidth]{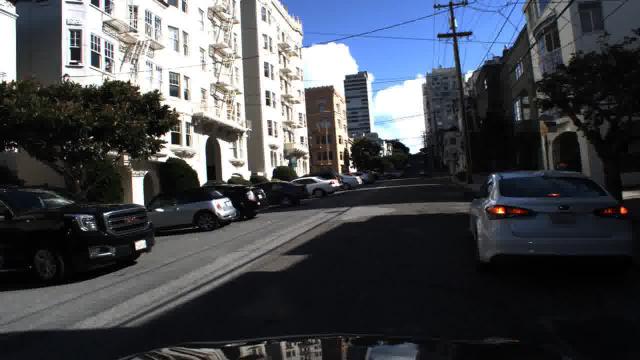}
			\includegraphics[width=.19\linewidth]{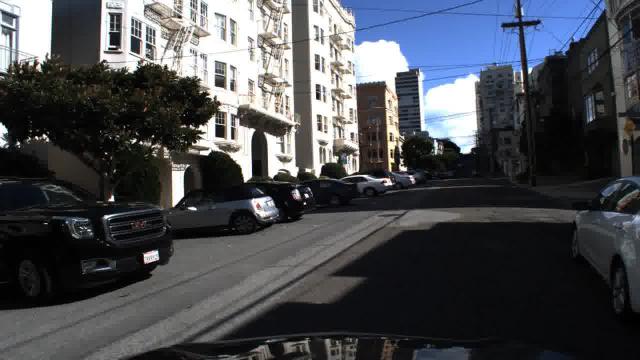}
			\includegraphics[width=.19\linewidth]{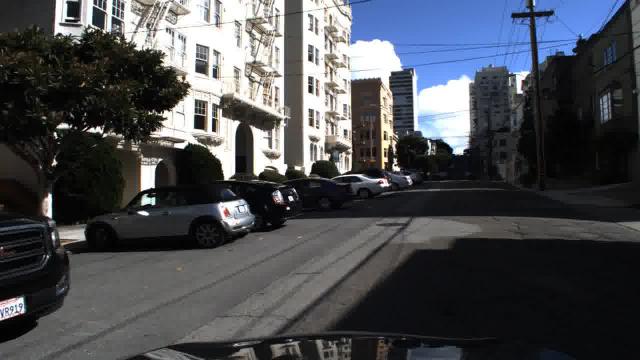}
			\includegraphics[width=.19\linewidth]{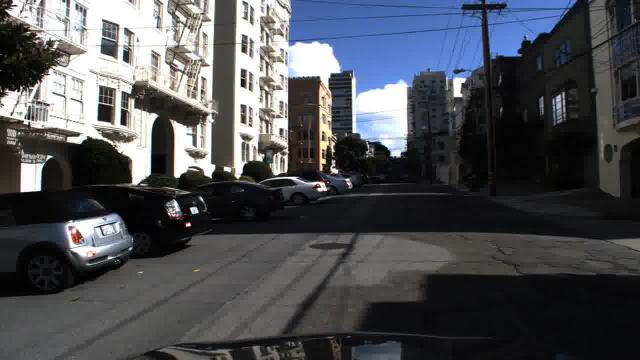}}
	\end{subfigure}
	\begin{subfigure}{1.0\textwidth}
		\centering
		\setlength{\fboxsep}{0pt}%
		\setlength{\fboxrule}{2pt}%
		\fcolorbox{green}{white}{\includegraphics[width=.19\linewidth]{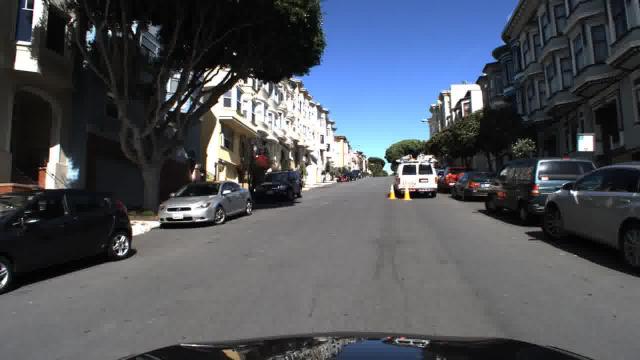}
			\includegraphics[width=.19\linewidth]{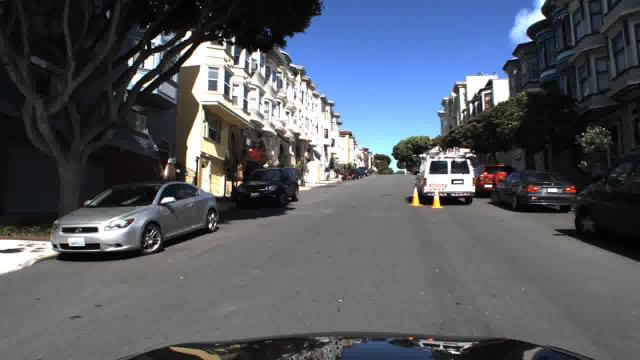}
			\includegraphics[width=.19\linewidth]{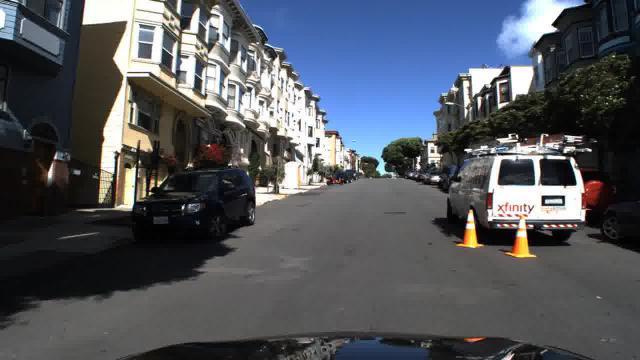}
			\includegraphics[width=.19\linewidth]{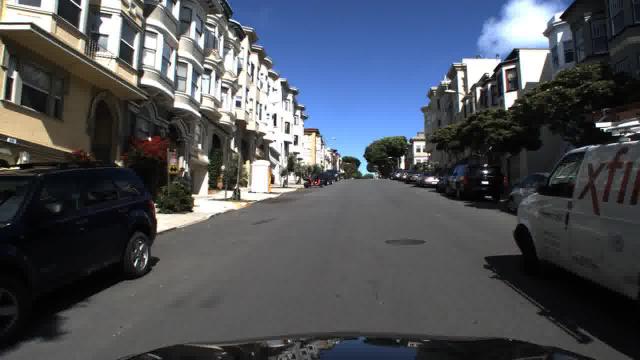}
			\includegraphics[width=.19\linewidth]{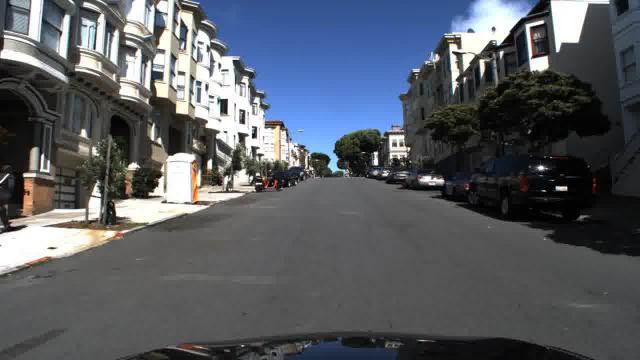}}
	\end{subfigure}
	\begin{subfigure}{1.0\textwidth}
		\centering
		\setlength{\fboxsep}{0pt}%
		\setlength{\fboxrule}{2pt}%
		\fcolorbox{green}{white}{\includegraphics[width=.19\linewidth]{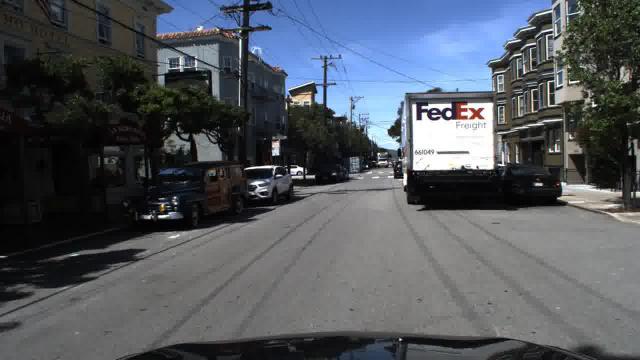}
			\includegraphics[width=.19\linewidth]{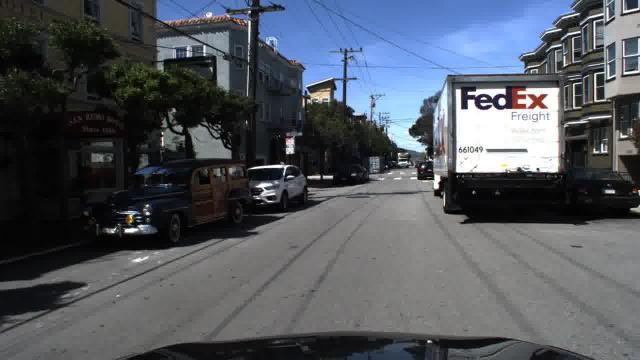}
			\includegraphics[width=.19\linewidth]{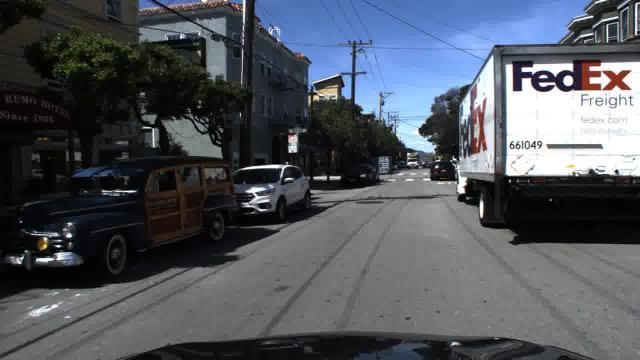}
			\includegraphics[width=.19\linewidth]{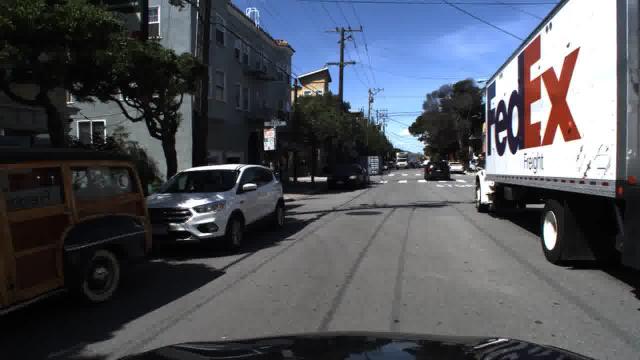}
			\includegraphics[width=.19\linewidth]{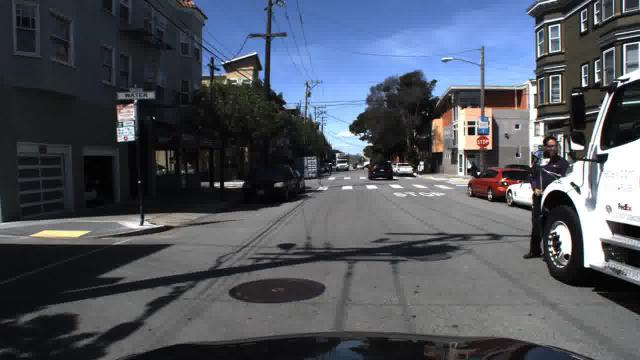}}
	\end{subfigure}

	\begin{subfigure}{1.0\textwidth}
		\centering
		\setlength{\fboxsep}{0pt}%
		\setlength{\fboxrule}{2pt}%
		
		\fcolorbox{blue}{white}{\includegraphics[width=.19\linewidth]{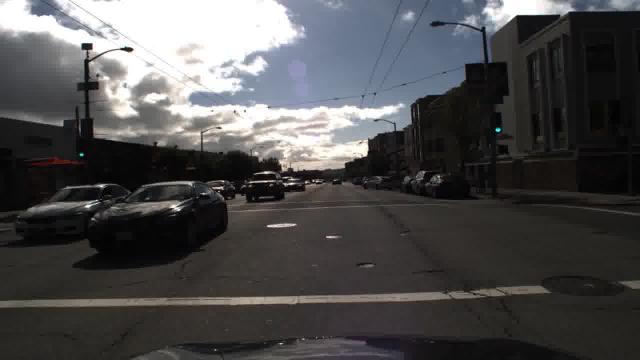}
			\includegraphics[width=.19\linewidth]{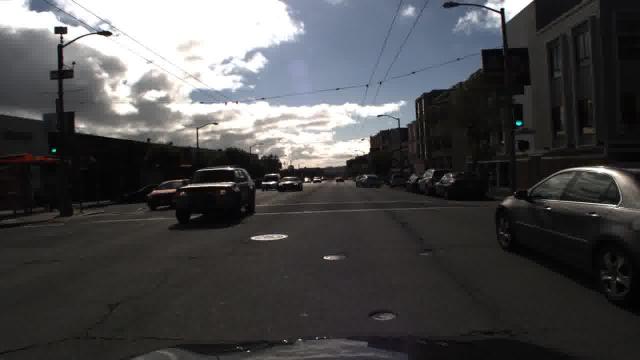}
			\includegraphics[width=.19\linewidth]{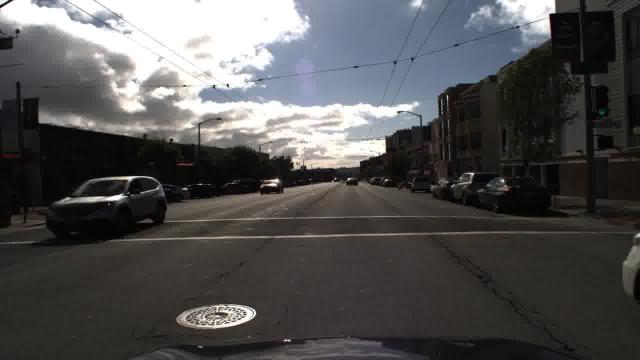}
			\includegraphics[width=.19\linewidth]{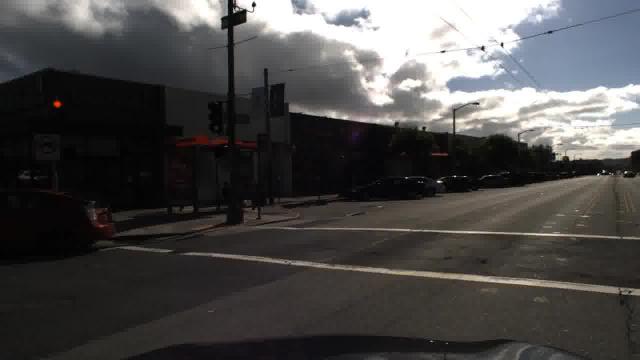}
			\includegraphics[width=.19\linewidth]{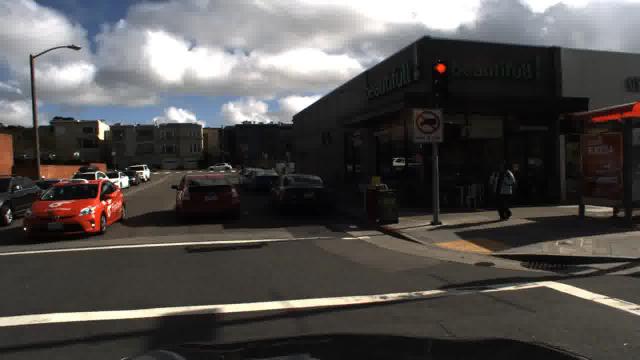}}
	\end{subfigure}
	\begin{subfigure}{1.0\textwidth}
		\centering
		\setlength{\fboxsep}{0pt}%
		\setlength{\fboxrule}{2pt}%
		\fcolorbox{green}{white}{\includegraphics[width=.19\linewidth]{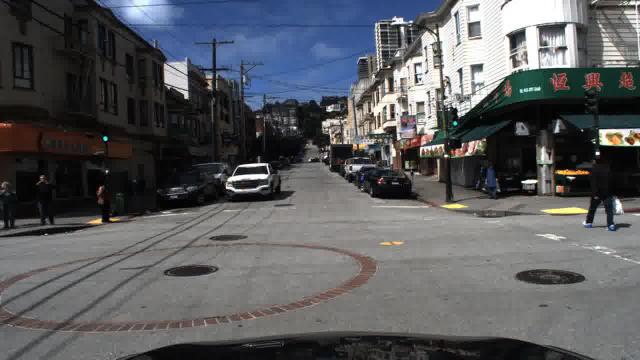}
			\includegraphics[width=.19\linewidth]{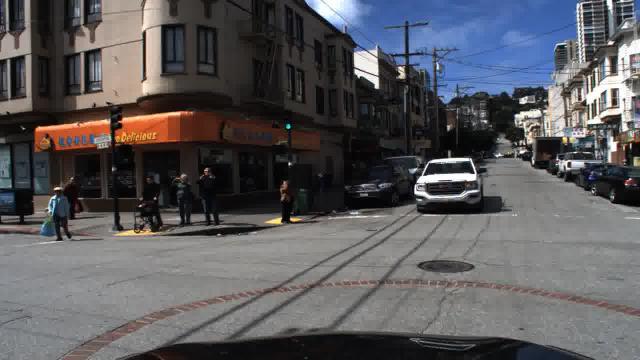}
			\includegraphics[width=.19\linewidth]{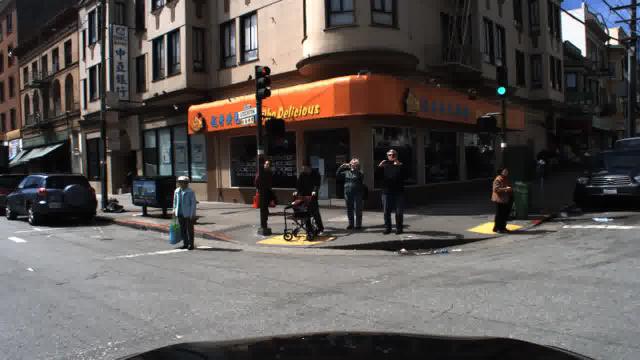}
			\includegraphics[width=.19\linewidth]{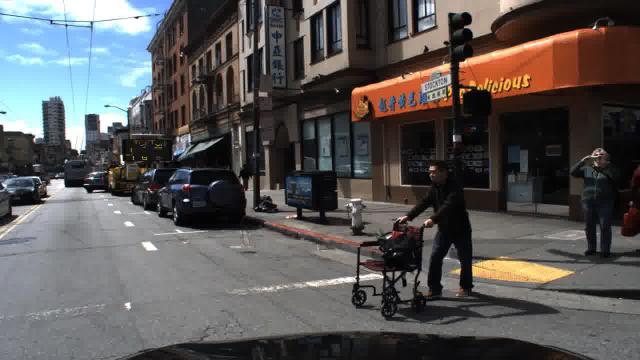}
			\includegraphics[width=.19\linewidth]{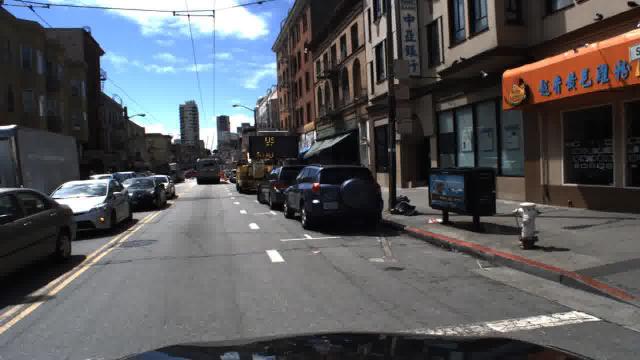}}
	\end{subfigure}
	\begin{subfigure}{1.0\textwidth}
		\centering
		\setlength{\fboxsep}{0pt}%
		\setlength{\fboxrule}{2pt}%
		\fcolorbox{green}{white}{\includegraphics[width=.19\linewidth]{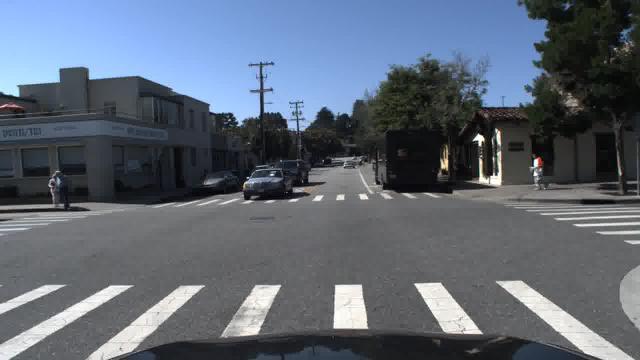}
			\includegraphics[width=.19\linewidth]{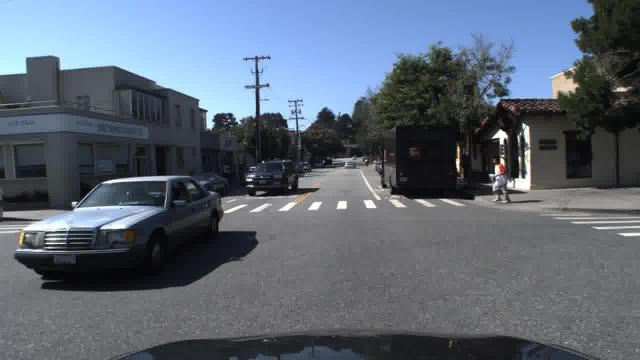}
			\includegraphics[width=.19\linewidth]{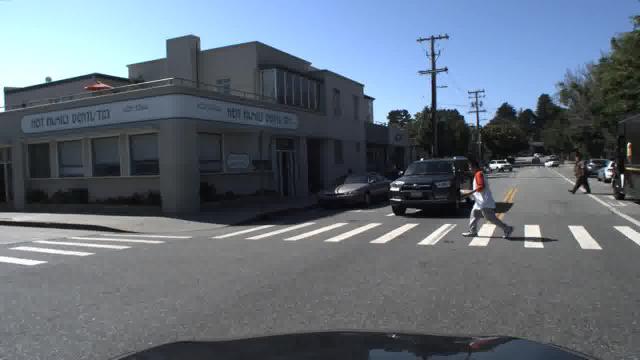}
			\includegraphics[width=.19\linewidth]{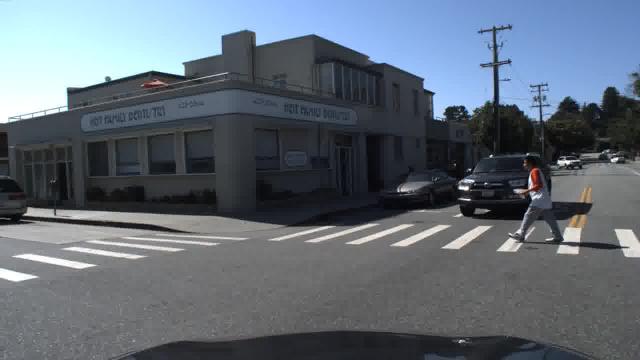}
			\includegraphics[width=.19\linewidth]{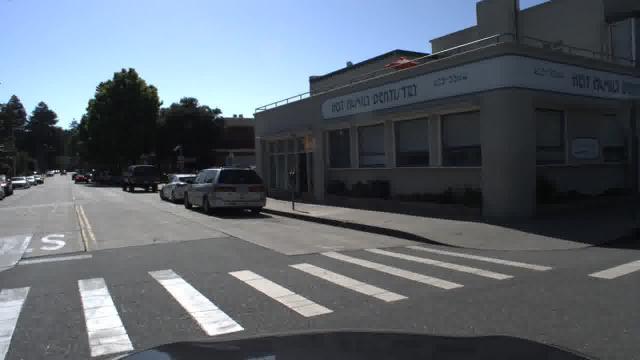}}
	\end{subfigure}
	
	\caption{Qualitative evaluation using two query events highlighted in blue. Every query is followed by the top two retrieval results highlighted in green. These queries emphasize the interaction between the ego-motion car and external environment factors. The first query shows the ego-motion car avoiding a parked car. The second query illustrates a \textit{left turn} maneuver interfered by pedestrian, cyclist and opposite traffic. These images are best viewed in color/screen.}
	\label{fig:quality_eval}
\end{figure*}


\subsection{Discussion}

Figure~\ref{fig:quality_eval} shows qualitative results for two queries: stimulus-driven and goal-oriented actions.
For each query, the top two retrieval results are highlighted in green. Beyond retrieving correct results, a few observations are worth noting. First, the top results show robustness to lighting conditions. Second, the learning embedding preserves the environment (street) style and layout as shown in the first query's top result. 

In the second query, the driver is obstructed by opposite traffic and pedestrians leading to a slow and more careful maneuver -- an important scenario for autonomous driving. While the current annotations provide no detailed sub-categorization for the \textit{left-turn} events, the results reveal a promising embedding. They demonstrate a similar slow maneuver to avoid opposite traffic, cyclist and pedestrian. A further qualitative study is required to support these observations.

One fundamental extension for this work is learning a joint embedding that fuses similarity across labels. For example, if the dataset annotations categorize goal-oriented actions like \textit{left} or \textit{right turn}, and independently categorize stimulus-driven actions like \textit{avoid a pedestrian} or \textit{stop for a traffic light}, is it possible to learn an embedding where \textit{left turn} maneuvers interfered by pedestrian are closer to each other than those interfered by traffic congestion? This would reduce the required amount of collected and labeled data, and cost accordingly. 








\section{Conclusion}
We present a re-formulation of triplet loss as a regression loss. This enables epistemic uncertainty to be employed in a retrieval context. Similar ranking losses can be cast to explore uncertainty and boost retrieval efficiency. We quantitatively evaluate our formulation in two domains. For person re-identification, comparable state-of-the-art results are achieved on Market-1501 and DukeMTMC-reID datasets. For autonomous navigation, heterogeneous modalities have various uncertainty levels for different similarities. Hence, we presented a multi-modal conditional retrieval system to fuse modalities and disentangle a per-similarity representation. It illustrates joint learning benefits in terms of lower computational cost and better performance. By employing Monte Carlo (MC) sampling, superior results are achieved. Quantitative evaluation emphasizes MC utility in a high uncertainty environment.

{\small
\bibliographystyle{ieee}
\bibliography{egbib}

\begin{thebibliography}{10}\itemsep=-1pt

\bibitem{arevalo2017gated}
J.~Arevalo, T.~Solorio, M.~Montes-y G{\'o}mez, and F.~A. Gonz{\'a}lez.
\newblock Gated multimodal units for information fusion.
\newblock {\em arXiv preprint arXiv:1702.01992}, 2017.

\bibitem{barbosa2018looking}
I.~B. Barbosa, M.~Cristani, B.~Caputo, A.~Rognhaugen, and T.~Theoharis.
\newblock Looking beyond appearances: Synthetic training data for deep cnns in
  re-identification.
\newblock {\em Computer Vision and Image Understanding}, 2018.

\bibitem{bishop2006pattern}
C.~M. Bishop.
\newblock Pattern recognition and machine learning (information science and
  statistics) springer-verlag new york.
\newblock {\em Inc. Secaucus, NJ, USA}, 2006.

\bibitem{caruana1997multitask}
R.~Caruana.
\newblock Multitask learning.
\newblock {\em Machine learning}, 1997.

\bibitem{chartsias2018multimodal}
A.~Chartsias, T.~Joyce, M.~V. Giuffrida, and S.~A. Tsaftaris.
\newblock Multimodal mr synthesis via modality-invariant latent representation.
\newblock {\em IEEE transactions on medical imaging}, 2018.

\bibitem{chen2017beyond}
W.~Chen, X.~Chen, J.~Zhang, and K.~Huang.
\newblock Beyond triplet loss: a deep quadruplet network for person
  re-identification.
\newblock In {\em CVPR}, 2017.

\bibitem{chen2017multi}
X.~Chen, H.~Ma, J.~Wan, B.~Li, and T.~Xia.
\newblock Multi-view 3d object detection network for autonomous driving.
\newblock In {\em CVPR}, 2017.

\bibitem{cheng2016person}
D.~Cheng, Y.~Gong, S.~Zhou, J.~Wang, and N.~Zheng.
\newblock Person re-identification by multi-channel parts-based cnn with
  improved triplet loss function.
\newblock In {\em CVPR}, 2016.

\bibitem{cho2014multi}
H.~Cho, Y.-W. Seo, B.~V. Kumar, and R.~R. Rajkumar.
\newblock A multi-sensor fusion system for moving object detection and tracking
  in urban driving environments.
\newblock In {\em Robotics and Automation (ICRA), 2014 IEEE International
  Conference on}, 2014.

\bibitem{dai2017temporal}
X.~Dai, B.~Singh, G.~Zhang, L.~S. Davis, and Y.~Q. Chen.
\newblock Temporal context network for activity localization in videos.
\newblock In {\em (ICCV)}, 2017.

\bibitem{feng2018leveraging}
D.~Feng, L.~Rosenbaum, F.~Timm, and K.~Dietmayer.
\newblock Leveraging heteroscedastic aleatoric uncertainties for robust
  real-time lidar 3d object detection.
\newblock {\em arXiv preprint arXiv:1809.05590}, 2018.

\bibitem{fukui2016multimodal}
A.~Fukui, D.~H. Park, D.~Yang, A.~Rohrbach, T.~Darrell, and M.~Rohrbach.
\newblock Multimodal compact bilinear pooling for visual question answering and
  visual grounding.
\newblock {\em arXiv preprint arXiv:1606.01847}, 2016.

\bibitem{funahashi1993approximation}
K.-i. Funahashi and Y.~Nakamura.
\newblock Approximation of dynamical systems by continuous time recurrent
  neural networks.
\newblock {\em Neural networks}, 1993.

\bibitem{gal2015bayesian}
Y.~Gal and Z.~Ghahramani.
\newblock Bayesian convolutional neural networks with bernoulli approximate
  variational inference.
\newblock {\em arXiv preprint arXiv:1506.02158}, 2015.

\bibitem{gal2015dropout}
Y.~Gal and Z.~Ghahramani.
\newblock Dropout as a bayesian approximation.
\newblock {\em arXiv preprint arXiv:1506.02157}, 2015.

\bibitem{gal2016dropout}
Y.~Gal and Z.~Ghahramani.
\newblock Dropout as a bayesian approximation: Representing model uncertainty
  in deep learning.
\newblock In {\em ICML}, 2016.

\bibitem{havaei2016hemis}
M.~Havaei, N.~Guizard, N.~Chapados, and Y.~Bengio.
\newblock Hemis: Hetero-modal image segmentation.
\newblock In {\em MICCAI}, 2016.

\bibitem{he2016deep}
K.~He, X.~Zhang, S.~Ren, and J.~Sun.
\newblock Deep residual learning for image recognition.
\newblock In {\em CVPR}, 2016.

\bibitem{hermans2017defense}
A.~Hermans, L.~Beyer, and B.~Leibe.
\newblock In defense of the triplet loss for person re-identification.
\newblock {\em arXiv preprint arXiv:1703.07737}, 2017.

\bibitem{hochreiter1997long}
S.~Hochreiter and J.~Schmidhuber.
\newblock Long short-term memory.
\newblock {\em Neural computation}, 1997.

\bibitem{huang2016learning}
C.~Huang, Y.~Li, C.~Change~Loy, and X.~Tang.
\newblock Learning deep representation for imbalanced classification.
\newblock In {\em CVPR}, 2016.

\bibitem{huang2016local}
C.~Huang, C.~C. Loy, and X.~Tang.
\newblock Local similarity-aware deep feature embedding.
\newblock In {\em NIPS}, 2016.

\bibitem{huang2017densely}
G.~Huang, Z.~Liu, L.~Van Der~Maaten, and K.~Q. Weinberger.
\newblock Densely connected convolutional networks.
\newblock In {\em CVPR}, 2017.

\bibitem{huang2018efficient}
P.-Y. Huang, W.-T. Hsu, C.-Y. Chiu, T.-F. Wu, and M.~Sun.
\newblock Efficient uncertainty estimation for semantic segmentation in videos.
\newblock {\em arXiv preprint arXiv:1807.11037}, 2018.

\bibitem{ilg2018uncertainty}
E.~Ilg, O.~Ci{\c{c}}ek, S.~Galesso, A.~Klein, O.~Makansi, F.~Hutter, and
  T.~Brox.
\newblock Uncertainty estimates and multi-hypotheses networks for optical flow.
\newblock In {\em ECCV}, 2018.

\bibitem{kaiser2017one}
L.~Kaiser, A.~N. Gomez, N.~Shazeer, A.~Vaswani, N.~Parmar, L.~Jones, and
  J.~Uszkoreit.
\newblock One model to learn them all.
\newblock {\em arXiv preprint arXiv:1706.05137}, 2017.

\bibitem{kendall2015bayesian}
A.~Kendall, V.~Badrinarayanan, and R.~Cipolla.
\newblock Bayesian segnet: Model uncertainty in deep convolutional
  encoder-decoder architectures for scene understanding.
\newblock {\em arXiv preprint arXiv:1511.02680}, 2015.

\bibitem{kendall2017uncertainties}
A.~Kendall and Y.~Gal.
\newblock What uncertainties do we need in bayesian deep learning for computer
  vision?
\newblock In {\em NIPS}, 2017.

\bibitem{kendall2017multi}
A.~Kendall, Y.~Gal, and R.~Cipolla.
\newblock Multi-task learning using uncertainty to weigh losses for scene
  geometry and semantics.
\newblock {\em arXiv preprint arXiv:1705.07115}, 3, 2017.

\bibitem{kingma2014adam}
D.~P. Kingma and J.~Ba.
\newblock Adam: A method for stochastic optimization.
\newblock {\em arXiv preprint arXiv:1412.6980}, 2014.

\bibitem{kokkinos2017ubernet}
I.~Kokkinos.
\newblock Ubernet: Training a universal convolutional neural network for low-,
  mid-, and high-level vision using diverse datasets and limited memory.
\newblock In {\em CVPR}, 2017.

\bibitem{li2017improving}
Y.~Li, Y.~Song, and J.~Luo.
\newblock Improving pairwise ranking for multi-label image classification.
\newblock In {\em CVPR}, 2017.

\bibitem{luo2018fast}
W.~Luo, B.~Yang, and R.~Urtasun.
\newblock Fast and furious: Real time end-to-end 3d detection, tracking and
  motion forecasting with a single convolutional net.
\newblock In {\em CVPR}, 2018.

\bibitem{nair2018exploring}
T.~Nair, D.~Precup, D.~L. Arnold, and T.~Arbel.
\newblock Exploring uncertainty measures in deep networks for multiple
  sclerosis lesion detection and segmentation.
\newblock In {\em MICCAI}, 2018.

\bibitem{ramanishka2018toward}
V.~Ramanishka, Y.-T. Chen, T.~Misu, and K.~Saenko.
\newblock Toward driving scene understanding: A dataset for learning driver
  behavior and causal reasoning.
\newblock In {\em CVPR}, 2018.

\bibitem{rasmussen2004gaussian}
C.~E. Rasmussen.
\newblock Gaussian processes in machine learning.
\newblock In {\em Advanced lectures on machine learning}. 2004.

\bibitem{ristani2016performance}
E.~Ristani, F.~Solera, R.~Zou, R.~Cucchiara, and C.~Tomasi.
\newblock Performance measures and a data set for multi-target, multi-camera
  tracking.
\newblock In {\em ECCV}, 2016.

\bibitem{ristani2018features}
E.~Ristani and C.~Tomasi.
\newblock Features for multi-target multi-camera tracking and
  re-identification.
\newblock {\em arXiv preprint arXiv:1803.10859}, 2018.

\bibitem{sankaranarayanan2016triplet}
S.~Sankaranarayanan, A.~Alavi, C.~Castillo, and R.~Chellappa.
\newblock Triplet probabilistic embedding for face verification and clustering.
\newblock {\em arXiv preprint arXiv:1604.05417}, 2016.

\bibitem{schroff2015facenet}
F.~Schroff, D.~Kalenichenko, and J.~Philbin.
\newblock Facenet: A unified embedding for face recognition and clustering.
\newblock In {\em CVPR}, 2015.

\bibitem{simonyan2014two}
K.~Simonyan and A.~Zisserman.
\newblock Two-stream convolutional networks for action recognition in videos.
\newblock In {\em NIPS}, 2014.

\bibitem{srivastava2014dropout}
N.~Srivastava, G.~Hinton, A.~Krizhevsky, I.~Sutskever, and R.~Salakhutdinov.
\newblock Dropout: a simple way to prevent neural networks from overfitting.
\newblock {\em The Journal of Machine Learning Research}, 2014.

\bibitem{su2016deep}
C.~Su, S.~Zhang, J.~Xing, W.~Gao, and Q.~Tian.
\newblock Deep attributes driven multi-camera person re-identification.
\newblock In {\em ECCV}, 2016.

\bibitem{sun2017svdnet}
Y.~Sun, L.~Zheng, W.~Deng, and S.~Wang.
\newblock Svdnet for pedestrian retrieval.
\newblock {\em arXiv preprint}, 2017.

\bibitem{szegedy2017inception}
C.~Szegedy, S.~Ioffe, V.~Vanhoucke, and A.~A. Alemi.
\newblock Inception-v4, inception-resnet and the impact of residual connections
  on learning.
\newblock In {\em AAAI}, 2017.

\bibitem{taha2018two}
A.~Taha, M.~Meshry, X.~Yang, Y.-T. Chen, and L.~Davis.
\newblock Two stream self-supervised learning for action recognition.
\newblock {\em DeepVision Computer Vision and Pattern Recognition Workshop},
  2018.

\bibitem{veit2017conditional}
A.~Veit, S.~J. Belongie, and T.~Karaletsos.
\newblock Conditional similarity networks.
\newblock In {\em CVPR}, 2017.

\bibitem{yu2015fast}
G.~Yu and J.~Yuan.
\newblock Fast action proposals for human action detection and search.
\newblock In {\em CVPR}, 2015.

\bibitem{zheng2015scalable}
L.~Zheng, L.~Shen, L.~Tian, S.~Wang, J.~Wang, and Q.~Tian.
\newblock Scalable person re-identification: A benchmark.
\newblock In {\em ICCV}, 2015.

\bibitem{zheng2016person}
L.~Zheng, Y.~Yang, and A.~G. Hauptmann.
\newblock Person re-identification: Past, present and future.
\newblock {\em arXiv preprint arXiv:1610.02984}, 2016.

\bibitem{zheng2017unlabeled}
Z.~Zheng, L.~Zheng, and Y.~Yang.
\newblock Unlabeled samples generated by gan improve the person
  re-identification baseline in vitro.
\newblock {\em arXiv preprint arXiv:1701.07717}, 2017.

\bibitem{zheng2018pedestrian}
Z.~Zheng, L.~Zheng, and Y.~Yang.
\newblock Pedestrian alignment network for large-scale person
  re-identification.
\newblock {\em IEEE Transactions on Circuits and Systems for Video Technology},
  2018.

\bibitem{zhou2017point}
S.~Zhou, J.~Wang, J.~Wang, Y.~Gong, and N.~Zheng.
\newblock Point to set similarity based deep feature learning for person
  reidentification.
\newblock In {\em CVPR}, 2017.

\end{thebibliography}
}

\clearpage

\end{document}